\documentclass{article}

\usepackage{arxiv,natbib}

\usepackage[utf8]{inputenc} 
\usepackage[T1]{fontenc}    
\usepackage{hyperref}       
\usepackage{url}            
\usepackage{booktabs}       
\usepackage{amsfonts}       
\usepackage{nicefrac}       
\usepackage{microtype}      
\usepackage{lipsum}
\usepackage{graphicx}
\usepackage{amsmath}
\usepackage{mathtools} 
\usepackage{latexsym}

\title{Unsupervised Spoken Utterance Classification}

\author{
Shahab Jalalvand \\
Senior Research Scientist \\
Interactions Corp. \\
Murray Hill, NJ, USA \\
sjalalvand@interactions.com \\
\And
Srinivas Bangalore \\
Director AI Research \\
Interactions Corp. \\
Murray Hill, NJ, USA \\
sbangalore@interactions.com \\
}

\begin{document}
\maketitle

\begin{abstract}
Intelligent virtual assistant (IVA) enables effortless conversations in call routing through spoken utterance classification (SUC) which is a special form of spoken language understanding (SLU).
Building an SUC system requires a large amount of supervised in-domain  data that is not always available.
In this paper, we introduce an unsupervised spoken utterance classification approach (USUC) that does not require any in-domain data except for the intent labels and a few para-phrases per intent.
USUC is consisting of a KNN classifier (K=1) and a complex embedding model trained on a large amount of unsupervised customer service corpus.
Among all embedding models, we demonstrate that Elmo works best for USUC.
However, an Elmo model is too slow to be used at run-time for call routing.
To resolve this issue, first we compute the uni- and bi-gram embedding vectors offline and we build a lookup table of n-grams and their corresponding embedding vector.
Then we use this table to compute sentence embedding vectors at run-time, along with back-off techniques for unseen n-grams.
Experiments show that USUC outperforms the traditional utterance classification methods by reducing the classification error rate from 32.9\% to 27.0\% without requiring supervised data.
Moreover, our lookup and back-off technique increases the processing speed from 16 utterance per second to 118 utterance per second.
\end{abstract}

\section{Introduction}

Intelligent virtual assistant (IVA) enables effortless conversations in call routing through spoken utterance classification (SUC) which is a special form of spoken language understanding (SLU) \cite{gorin1997how,tur2011spoken}.
Many companies in different marketing domains such as finance, retail and hospitality automate their call centers through SUC systems to reduce the human resources.
Despite notable achievements, there are still several issues such as channel and domain variation that prevent complete user satisfaction.

\begin{figure}[h]
\centering
\includegraphics[trim=0cm 0cm 0cm 0cm, clip=true, width=\linewidth]{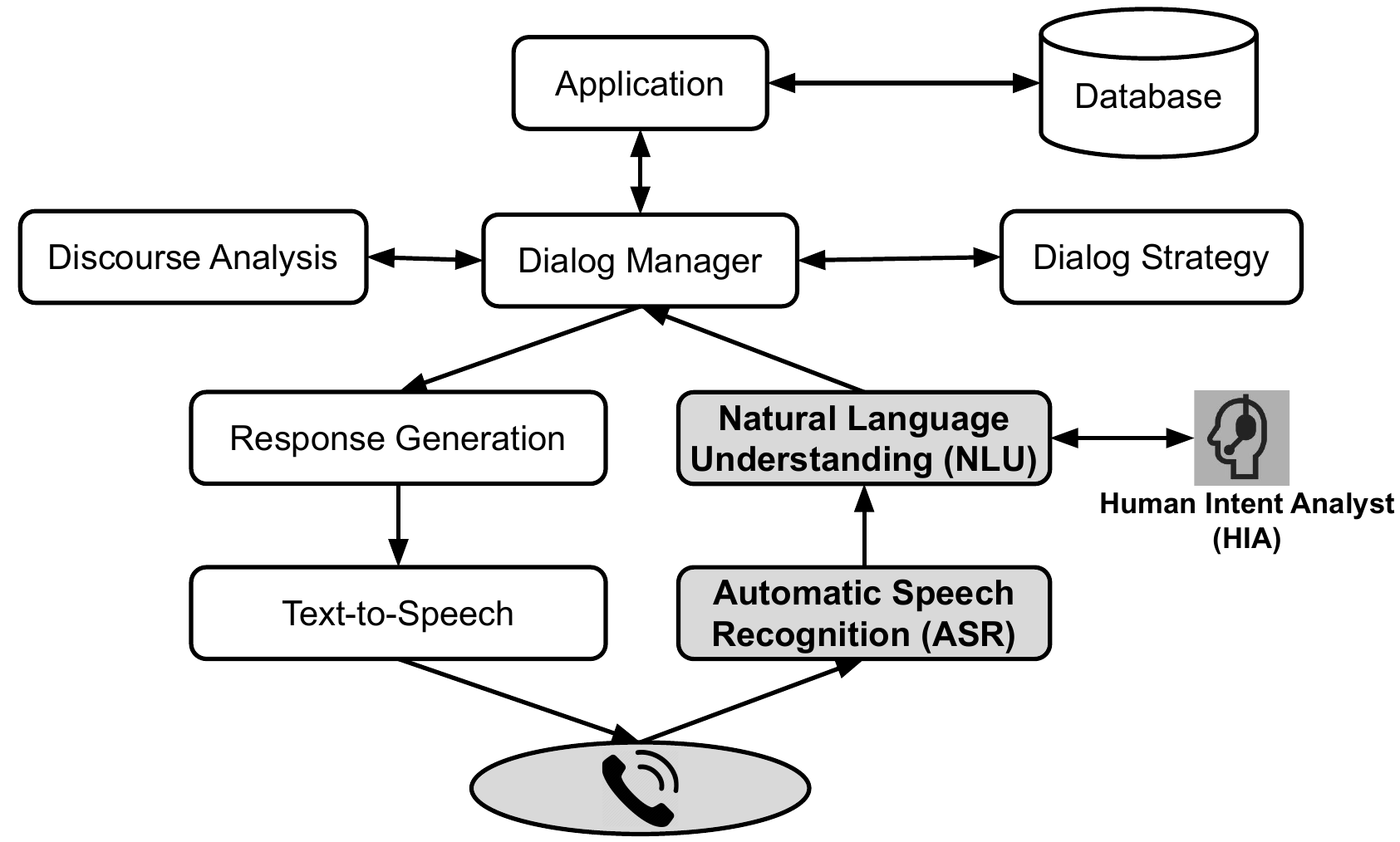}
\caption{Call routing system with human in the loop.}
\label{fig:slu}
\end{figure}

Figure \ref{fig:slu} shows a typical SUC system embedded in a call routing architecture \cite{gorin1997how,huang2001spoken,busayapongchai2012methods}.
A transaction starts with a prompt \textit{"How may I help you?"}.
The user's utterance is given to an automatic speech recognition (ASR) module to generate  text \textit{"i'd like to add a driver to my policy"}.
The text is transformed to a natural language understanding (NLU) module to recognize the intent \textit{"ADD\_DRIVER"}.
If the confidence score is not high enough, the call is routed to a human intent analyst (HIA) who determines the correct intent.
Based on the predicted intent, dialog manager provides the proper response and the dialog continues.

In contrast to ASR, NLU entails more detailed training with respect to the domain it is being designed for.
Because the callers' intents differ dramatically from one application to the other. 
Therefore the NLU models require a large amount of supervised in-domain data. 
\citet{dauphin2013zero} propose a DNN-based approach that can learn discriminative semantic features without supervision.
To overcome the lack of data, \citet{shahab2018automatic} proposed a data expansion algorithm  through identifying the discriminative phrases and selecting the samples that contain those phrases from a data pool.

In this paper we introduce an unsupervised spoken utterance classification (USUC) that does not require any in-domain data except for a list of intent labels (classes) and a few para-phrases per intent.
USUC aims to map the ASR hypothesis into one of the para=phrases through computing cosine similarities in an embedding space.
To do this, USUC is consisting of a simple KNN (K=1) classifier and a complex embedding model.
We demonstrate that for this task Elmo \cite{Peters2018deep} works better than the other embedding approaches.
However, Elmo is too slow to be used at run-time.
Industrial call routing systems cannot tolerate such a delay to answer the user.
To resolve this issue, we compute the embedding vectors of all possible n-grams and we use back-off techniques \cite{katz1987estimation} for unseen n-grams.
Experiments show 5.9\% absolute improvement in classification error rate with regard to the state-of-the-art approaches.
Moreover, our lookup technique increases the utterance processing speed from 16 utterances per second to 67 utterance.

The proposed USUC system is indeed inspired from human's behavior (HIA in fig. \ref{fig:slu}).
HIA agents are asked to memorize the intent labels and the para-phrases.
When they listen to an utterance they do either a) key-word spotting if the para-phrase is spoken in the utterance or b) semantic analysis to find the closest para-phrase.
In USUC we replace the HIA's semantic knowledge with an embedding model that is trained on a large unsupervised corpus from various customer service applications.
To the best of our knowledge, this work is the first attempt to simulate human's behavior for detecting the caller's intent.

The rest of this paper is organized as follows.
In Section 2, we briefly describe the SUC system;
Section 3 describes USUC;
Section 4 contains the experiments and results;
and Section 5 concludes the paper.

\section{Spoken Utterance Classification (SUC)}
In an SUC system, a speech recognition module finds $\hat{W}$ which is a word sequence that maximizes the posterior probability given utterance $X$ and then
a language understanding module finds the intent class $\hat{C}$ that maximizes posterior probability given $\hat{W}$.

\begin{flalign}
\label{eq:slu}
& ASR: \hat{W} = \underset{W} {argmax} \{P(W|X)\} \\
& NLU: \hat{C} = \underset{C} {argmax} \{P(C|\hat{W})\}.
\end{flalign}

In this paper, we focus on NLU and we assume that the ASR system is previously built and it is fixed.

To predict $\hat{C}$ from hypothesis $\hat{W}$ different machine learning algorithms can be applied.
A widely used learning algorithm is support vector machine (SVM) \cite{cortes1995support} with $L2$ feature normalization.
The feature set is a bag of n-grams (typically $n=1,2$).
The dimension of the feature vector is equal to the total number of uni- and bi-grams in the training data.
$k$-th element of the vector is $1$, if its corresponding n-gram is seen in $\hat{W}$.

Other approaches are based on deep neural networks (DNN) \cite{tur2012towards}.
The input is typically a sequence of words. 
These words are represented with word embeddings, character embeddings, or both. 
Sentences are represented as a combination of word representations by pooling the final output of a recurrent neural network.
A final  softmax layer is used for classification \cite{kim2014convolutional}. 
Our DNN toolkit \cite{pressel2018baseline} supports these models as well as an MLP built on pre-trained word embeddings with maxover-time pooling or mean-pooling.

Both aforementioned methods require a large amount of supervised in-domain data and they are strongly dependent on distribution of labels.

\section{Unsupervised Spoken Utterance Classification (USUC)}

USUC is consisting of a KNN classifier (K=1) and an embedding model.
The embedding model is used to compute an embedding vector for the ASR hypothesis $\hat{W}$ and for all the para-phrases.
The KNN classifier finds the closest para-phrase to $\hat{W}$, based on cosine similarity measure.
Considering K=1 in the KNN classifier allows the model to be less dependent on distribution of intent labels.
Each para-phrase is a side information and it is correspondent to one intent label which is the predicted label by USUC.

Assume that there are $M$ intent classes $\{C^1,...,C^M\}$.
For each class $C^i$ there are $P^i$ para-phrases $\{p_1^i,...,p_{P^i}^i\}$.
Given $\hat{W}$, USUC finds the intent class $\hat{C}=C^I$ :

\begin{flalign}
\label{eq:usuc}
& (I,J) = \\ \nonumber
& \underset{(i,j)} {argmax} \{cosim(Elmo(\hat{W}),Elmo(p^i_j)): i \in [1,M] , j \in [1,P^i] \}.
\end{flalign}

\noindent
Function $Elmo(W)$ returns the embedding vector of sentence $W$ and $cosim()$ computes the cosine similarity between two vectors. 

\subsection{Embedding model}
Many approaches have been developed to produce word/sentence embeddings such as neural probabilistic language model \cite{bengio2003a}, Word2Vec \cite{mikolov2013distributed}, GloVe \cite{pennington2014glove} and more recently InferSent \cite{conneau2017supervised} and Elmo \cite{Peters2018deep}.
Each method has pros and cons and it's not trivial to find one that works consistently well in different downstream tasks \cite{christian2018evaluation}. We demonstrate that for our USUC system, Elmo outperforms the others by a large margin.

In Elmo, the embeddings are computed from the internal states of a two-layers bidirectional Language Model (LM).
The inputs are characters rather than words, thus this model can compute meaningful representations even for out-of-vocabulary words.
However, Elmo is context-dependent and it needs to be activated individually for each utterance.
This is not feasible in industrial call routing systems because it dramatically increases the response time.

\subsection{Embedding look-up table}
\label{sec:lookup}
To exploit Elmo embeddings at run-time, we compute the embedding vectors for words and n-grams offline and we store them in a lookup table.
These n-grams are extracted from a generic text corpus described in Section \ref{sec:exp}.
To compute the embedding vector for $\hat{W}=\{w_1...w_N\}$, we use three methods:

\noindent
\textbf{Lookup Word:}
This table is consisting of words and their corresponding embedding vector $Elmo("w_i")$.
Sentence embedding vector is computes by average pooling across the words:

\begin{flalign}
\label{eq:lookupW}
& \hat{W}_{emb} = \frac{1}{N} \sum_{i=1}^{N}{Elmo("w_i")}.
\end{flalign}

\noindent
$Elmo(w)$ returns the embedding vector of the word $w$ when the word is passed to the Elmo model as a 1-word sentence.
Therefore, there is no context for computing the word embedding vectors.

\textbf{Lookup N-gram:}
This table is consisting of n-grams and their corresponding vectors.
Sentence vector is computed by average-pooling across the n-grams in the hypothesis

\begin{flalign}
\label{eq:lookupN}
& \hat{W}_{emb} = \frac{1}{N} \sum_{i=n}^{N}{emb("w_{i-n+1}^{i}")} \\ \nonumber
& emb("w_{i-n+1}^{i}") =\left\{\begin{matrix}
Elmo("w_{i-n+1}^{i}") \ \ \  if\ C("w_{i-n+1}^{i}") > 0 \\ \\
emb("w_{i-n+2}^{i}") \ \ \ \ \ \ \ \ \ \ \ \ \ \ \ \ \ \ \ \ \ \ \ \ \ \ \ \ else \\ 
\end{matrix}\right.
\end{flalign}

\noindent
$C(W)$ is the count of string $W$.
Basically in the above equation if an n-gram is found in the table, its embedding vector is used directly. 
Otherwise we shrink the context and we query the table.

\textbf{Lookup N-gram with back-off:}
In this table, for unseen n-grams we use a modified back-off technique \cite{katz1987estimation}.
\begin{flalign}
\label{eq:lookupNB}
& \hat{W}_{vec} = \frac{1}{N} \sum_{i=n}^{N}{vec("w_{i-n+1}^{i}")} \\ \nonumber
& vec("w_{i-n+1}^{i}") =\left\{\begin{matrix}
emb("w_{i-n+1}^{i}") \ \ \  if\ C("w_{i-n+1}^{i}") > 0 \\ \\
bof({w_{i-n+1}^{i-1}}) . vec("w_{i-n+2}^{i}")  \ \ else \\ 
\end{matrix}\right.
\end{flalign}

\noindent
$bof()$ is the back-off score of the context.
These scores are computed after training a back-off n-gram language model \cite{stolcke2002srilm} on the generic text corpus described in Section \ref{sec:exp}.

\section{Experiments}
\label{sec:exp}
Table \ref{tab:data} shows the statistics of the available datasets.
The first column represents the total number of utterances, the second last column is the total number of intent labels and the last column shows the total number of para-phrases in each set.

\begin{table}[h]
    \centering
    \begin{tabular}{l|c|c|c|c|c} 
    Dataset  & Sent. & Tok. & Vocab. & Intents & Para-phrases \\ \hline \hline
    Generic & 5.3M & 40M & 12.3K & 5.2K & N/A \\
    In-domain & 478 & 1.2K & 414 & 156 & 456 \\
    Expanded & 11.7K & 111K & 2.2K & 156 & N/A \\
    Test & 1.2K & 4.4K & 401 & 68 & N/A \\ \hline 
    \end{tabular}
    \caption{Statistics of the available datasets.}
    \label{tab:data}
\end{table}

\textbf{Generic} is consisting of 5.3 million utterances and 5.2 thousands unique intents.
This dataset covers 6 domains: communications, finance and banking, healthcare, retail and technology, travel and hospitality and utilities. 
\textbf{In-domain} is a list of intents and para-phrases in the insurance domain.
It contains 156 unique intents and 456 para-phrases, that is 2.9 para-phrases per intent on average.
This reflects the difficulty of the task in hand.
\textbf{Expanded} is obtained after data expansion using para-phrases as seeds and the generic corpus as the data pool \cite{shahab2018automatic}.
\textbf{Test} is composed of 1200 utterances with 68 unique intents which is also from insurance domain. 
All the intents in \textit{Test} are covered in \textit{In-domain}, however only 7.7\% of these intents are seen in \textit{Generic}. Table \ref{tab:intentexample} shows an example of the intents, para-phrases and real utterances.

\begin{table}[h]
\centering
\small
\begin{tabular}{p{1.8cm}|p{1.6cm}|p{4cm}}
Intents & Para-phrases & Utterances \\ \hline\hline    
ADD\_DRIVER & add driver & add a driver to my policy \\\hline
ADD\_DRIVER & add person & adding someone to my insurance\\\hline
CANCEL\_PAY & cancel payment & uh about canceling automatic payment\\\hline
FILING & filing & i need form insurance verification\\\hline
MOVE\_DUE & move due date & reschedule my due date  \\\hline
MOVE\_DUE & change payment date &  i wanna change my payment data \\\hline
\end{tabular}
\caption{In-domain data examples.}
\label{tab:intentexample}
\end{table}

Our goal is to enhance the classification error rate (CER) measure on the \textit{Test} set using \textit{In-domain} and \textit{Generic} sets.

\subsection{Automatic Speech Recognition}
The ASR system in this work is pre-trained and it does not change across different experiments.
This system consists of a 3-gram language model and hybrid DNN acoustic model trained with the cross-entropy criterion followed by the state-level Minimum Bayes Risk (sMBR) objective. 
The language model is trained on the \textit{Generic} set with 40 million words.
The acoustic model is trained on ~400 hours of customer utterances.
Decoding was performed with a wide beam setting yielding word error rate of 14.5\% on the \textit{Test} set.

\subsection{Terms of Comparison}
We compare five utterance classification methods:

\begin{itemize}
    \item \textbf{Human} shows the performance of the human intent analysts (HIA) who listen to the \textit{Test} utterances and detect the intents. Obviously this term is far better than the other machine learning terms;
    \item \textbf{SVM\_UC+In-domain} is an SVM trained on  \textit{In-domain};
    \item \textbf{SVM\_UC+Expanded} is an SVM trained on  \textit{Expanded};
    \item \textbf{DNN\_UC+Expanded} is a DNN model \cite{pressel2018baseline} trained on the \textit{Expanded} data;
    \item \textbf{USUC} is our proposed unsupervised method with different embedding models.
\end{itemize}

\subsection{Results}

\begin{table}[]
    \centering
    \begin{tabular}{l|c} 
    Term of Comparison  & CER\% \\ \hline \hline
    Human       & 12.7  \\ \hline
    SVM\_UC+In-domain       & 46.7  \\
    SVM\_UC+Expanded & 33.3 \\ 
    DNN\_UC+Expanded & 32.9 \\ \hline
    USUC+Word2Vec &  45.9 \\
    USUC+GloVe & 41.2 \\
    USUC+InferSent &  37.2 \\
    USUC+Elmo  & 27.0 \\ \hline
    \end{tabular}
    \caption{Classification error rate (CER) result of different intent predictors.}
    \label{tab:result}
\end{table}

Table \ref{tab:result} shows the classification error rate (CER) results on the \textit{Test} utterances.
\textit{Human} performs at 12.7\% CER. 
Note that one reason for the large gap between \textit{Human} and other machine learning algorithms is due to the ASR errors (14.5\% word error rate).
\textit{SVM\_UC+In-domain} yields 46.7\% CER, obviously because the training set is too small (456 samples for 156 labels).
\textit{SVM\_UC+expansion} improves the CER results by 13.4\% absolutely.
Thanks to the expansion approach \cite{shahab2018automatic}, we could expand the para-phrases from 456 samples to 11.7K using \textit{In-domain} as seed and \textit{Generic} as data pool.
\textit{DNN\_UC+Expanded} improves over the SVM models by 0.4\%.
Our DNN is a convolutional neural network (CNN) with 600 hidden neurons and the softmax layer is 156 dimensions \cite{pressel2018baseline}.
We also tried to expand more data and train a bigger DNN but it did improve the results, probably due to adding more noisy data.

The last set pf results shows the performance of different embedding approaches. 
First we use a Word2Vec \cite{mikolov2013distributed} model trained on \textit{Generic}.
We define the embedding vector of 100 dimensions, window size of 5 and minimum count as 1.
We use this model to obtain the word-level embedding vectors and then we do average-pooling to compute sentence-level vectors.
This approach is slightly better than our first baseline \textit{SVM\_UC}, though far worse than our state-of-the-art \textit{DNN+expansion}.
After that, we exploit GloVe-840b-300d \cite{pennington2014glove} as the embedding model and we got 4.7\% improvement over Word2Vec. 
Then, we apply InferSent \cite{conneau2017supervised} which is a sentence-level embedding model.
We use encoder InferSent version 2 with the default parameters.
This model yields 37.2\% CER which is better than GloVe but still worse than \textit{DNN+expansion}. 
Finally in \textit{USUC+Elmo} we use an Elmo model that is trained on the \textit{Generic} set.
The dimension of the word embedding vectors is 1024 and average pooling is used to compute sentence-level vectors.
This approach yields the best result 27.0\% CER which is significantly better than the state-of-the-art approach \textit{DNN+expansion} by 5.9\% absolute improvement.
Other pooling methods such as max-pool and p-mean did not show consistent improvement.

\subsection{Run-time Analysis}

\begin{table}[]
    \centering
    \begin{tabular}{l|c|c} 
    Method & CER & Speed \\
    & & (utterance/sec) \\\hline\hline
    USUC+Elmo & 27.0 & 16 \\ \hline
    USUC+Elmo+LookupWord & 29.7 & 120 \\
    USUC+Elmo+LookupNgram & 28.5 & 118 \\
    USUC+Elmo+LookupNgram+Backoff & 27.2 & 118  \\\hline 
    \end{tabular}
    \caption{Run-time speed analysis.}
    \label{tab:time}
\end{table}

We analyze the run-time speed of \textit{USUC+Elmo} approach with different lookup methods.
In Table \ref{tab:time}, the first row shows the CER and the computation speed of the original \textit{USUC+Elmo} approach, when the embedding model is activated individually for each test utterance.
As we see, the speed of this approach is quite low (16 utterances per second).
Then we use the lookup word embedding technique described in Section \ref{sec:lookup}. 
Basically we compute the embedding vector of each word offline and we store it in a lookup table.
The size of this table is $[12.3K \times 1204]$, being respectively the vocabulary size of the \textit{Generic} set by the embedding dimension. 
As we see this method is 7.5 times faster than the original approach, though it is 1.7\% worse in terms of accuracy.
Lookup n-gram enlarges the table to $[715K \times 1204]$, though it does not influence the process speed dramatically because we use the memory mapped files to store the tables.
The \textit{LookupNgram} approach improves the CER\% in compare to the word table.
Finally the best result is achieved by including the back-off scores in \textit{USUC+Elmo+LookupNgram+Backoff} (Eq. \ref{eq:lookupNB}).
We observe that this approach improves the performance of the lookup embedding techniques almost as accurate as the original approach \textit{USUC+Elmo}, though it's 7.3\% times faster.

\section{Conclusion}
We introduced a novel unsupervised spoken utterance classification approach for call routing systems, removing the need for large supervised in-domain data.
Inspired from human's behavior, our proposed approach utilized an embedding model as a semantic knowledge to detect the caller's intent.
We also proposed efficient techniques based on embedding lookup tables and back-off techniques to exploit complex embedding models at run-time.

\bibliographystyle{ACM-Reference-Format}
\bibliography{main}

\end{document}